%% file: Main.tex
\documentclass[11pt,a4paper]{article}
\pdfoutput=1
\usepackage{authblk}
\usepackage{emnlp2021}
\usepackage{times}
\usepackage{latexsym}
\usepackage{listings}
\usepackage{backnaur}
\usepackage{graphicx}
\usepackage{syntax} 
\usepackage{multirow}
\usepackage{booktabs}
\usepackage{xcolor}

\usepackage[linesnumbered,ruled,vlined]{algorithm2e}

\usepackage{listings}

\usepackage{microtype}

\newcommand\blfootnote[1]{%
  \begingroup
  \renewcommand\thefootnote{}\footnote{#1}%
  \addtocounter{footnote}{-1}%
  \endgroup
}

\title{Text2App: A Framework for Creating Android Apps from Text Descriptions}

\author[1*]{Masum Hasan}
\author[1*]{Kazi Sajeed Mehrab}
\author[2]{Wasi Uddin Ahmad}
\author[1]{Rifat Shahriyar}
\affil[1]{Bangladesh University of Engineering and Technology (BUET)}
\affil[2]{University of California, Los Angeles (UCLA)}
\affil[1]{\textit{masum@ra.cse.buet.ac.bd, 1505025.ksh@ugrad.cse.buet.ac.bd, rifat@cse.buet.ac.bd}}
\affil[2]{\textit{wasiahmad@cs.ucla.edu}
}

\date{}

\begin{document}
\maketitle


\begin{abstract}
\input{Sections/Abstract}
\blfootnote{* Equal contribution}
\end{abstract}

\section{Introduction}
\label{sec:introduction}

\input{Sections/Introduction}

\section{Related Works}
\label{sec:relatedworks}

\input{Sections/RelatedWorks}


\section{Text2App}
\label{sec:text2app}

\input{Sections/Text2App}

\section{Evaluation}
\label{sec:evaluation}

\input{Sections/Evaluation}

\section{Scope, Limitation, and Future Work}
\label{sec:future}

\input{Sections/DiscussionAndFuture}

\section{Conclusion}
\label{sec:conclusion}

\input{Sections/Conclusion}

\section*{Acknowledgement}
\label{sec:acknowledgement}
\input{Sections/Acknowledgement}

\bibliographystyle{aclnatbib}
\bibliography{Main}

 \appendix
 \clearpage
 \twocolumn[{%
  \centering
  \Large\bf Supplementary Material: Appendices \\ [20pt]
 }]
 \input{Sections/AppendixText}

\end{document}

%% file: Sections/Abstract.tex
We present Text2App -- a framework that allows users to create functional Android applications from natural language specifications. The conventional method of source code generation tries to generate source code directly, which is impractical for creating complex software. We overcome this limitation by transforming natural language into an abstract intermediate formal language representing an application with a substantially smaller number of tokens. The intermediate formal representation is then compiled into target source codes. This abstraction of programming details allows seq2seq networks to learn complex application structures with less overhead. In order to train sequence models, we introduce a data synthesis method grounded in a human survey. We demonstrate that Text2App generalizes well to unseen combination of app components and it is capable of handling noisy natural language instructions. We explore the possibility of creating applications from highly abstract instructions by coupling our system with  GPT-3 -- a large pretrained language model. We perform an extensive human evaluation and identify the capabilities and limitations of our system. The source code, a ready-to-run demo notebook, and a demo video are publicly available at \url{https://github.com/text2app/Text2App}.

%% file: Sections/Introduction.tex

\begin{figure}[h]
    \centering
    \includegraphics[width=\linewidth]{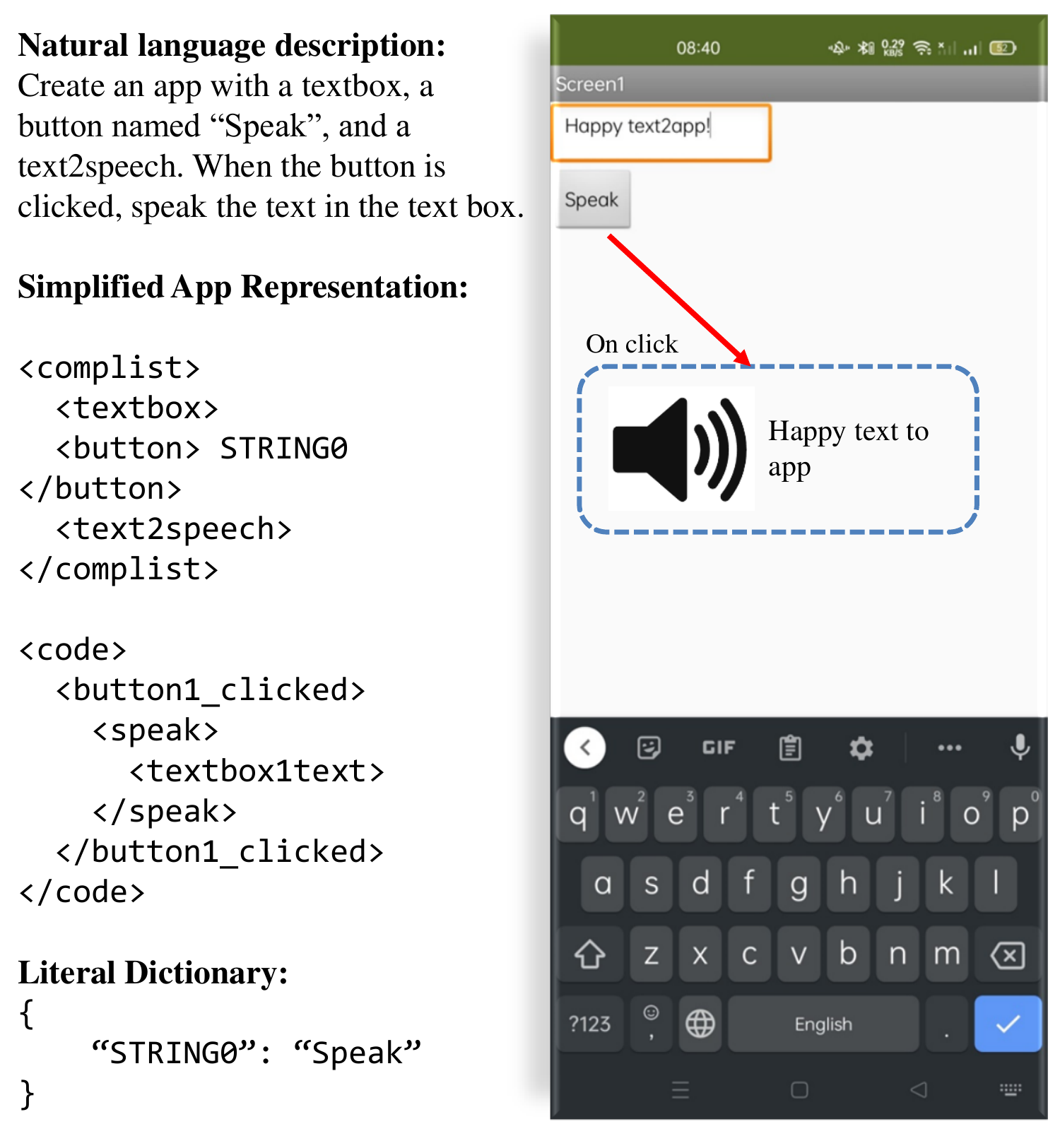}%
    \caption{An example app created by our system that speaks the textbox text on button press. The natural language is machine translated to a simpler intermediate formal language which is compiled into an app source code. Literals are separated before machine translation.}
    \label{fig:app}
\end{figure}

Mobile application developers often have to build applications from natural language requirements provided by their clients or managers. An automated tool to build functional applications from such natural language descriptions will significantly value this application development process. For many years, researchers have been trying to generate source code from natural language descriptions \cite{neubig,latent}, with the aspiration to automatically generate full-fledged software systems further down the road. To date, however, the task of source code generation has turned out to be highly difficult -- the best deep neural networks, consisting of hundreds of millions of parameters and trained with hundreds of gigabytes of data, fails to achieve an accuracy higher than 20\% \cite{plbert,codexglue}. Till date, the ambition to produce software automatically from natural language descriptions has remained a distant reality. 

In this work, we present Text2App, a novel pipeline to generate Android Mobile Applications (app) from natural language (NL) descriptions. Instead of an end-to-end learning-based model, we break down the challenging task of app development into modular components and apply learning-based methods only where necessary. We create a formal language named \textit{Simplified App Representation (SAR)}, to represent an app with a minimal number of tokens and train a sequence-to-sequence neural network to generate this formal representation from an NL. Fig. \ref{fig:app} shows an example of a formal representation created from a given NL. Using a custom-made compiler, we convert the simplified formal representation to the application source code from which a functional app can be built. We create a data synthesis method and a BERT-based NL augmentation method to synthesize realistic NL-SAR parallel corpus. 

We demonstrate that the compact app representation allows seq2seq models to generate app from significantly noisy input, even being able to predict combinations it has not seen during training. 
Moreover, we open source our implementation to the community, and lay down the groundwork to extend the features and functionalities of Text2App beyond what we demonstrated in this paper.  





%% file: Sections/RelatedWorks.tex
Historically, deep learning based program generation tended to focus on generating unit functions or methods from natural language instructions using sequence-to-sequence or sequence-to-tree architectures \cite{latent,neubig, brockschmidt2018generative,parisotto2016neurosymbolic,ast1,plbert,codexglue}.
The other type of works in program generation that sparked researcher's interest is generating GUI source code from a screenshot, hand-drawn image, or text description of the GUI \cite{pix2code,jain2019sketch2code,robinson2019sketch2code,gui1,gui2,text-gui}. These works are limited to generating GUI design only, and does not naturally extend to functionality based programming.
To the best of our knowledge, ours is the first work on developing working software with interdependent functional components from natural language description.

%% file: Sections/Text2App.tex
\begin{figure*}
    \centering
    \includegraphics[width=.8\linewidth]{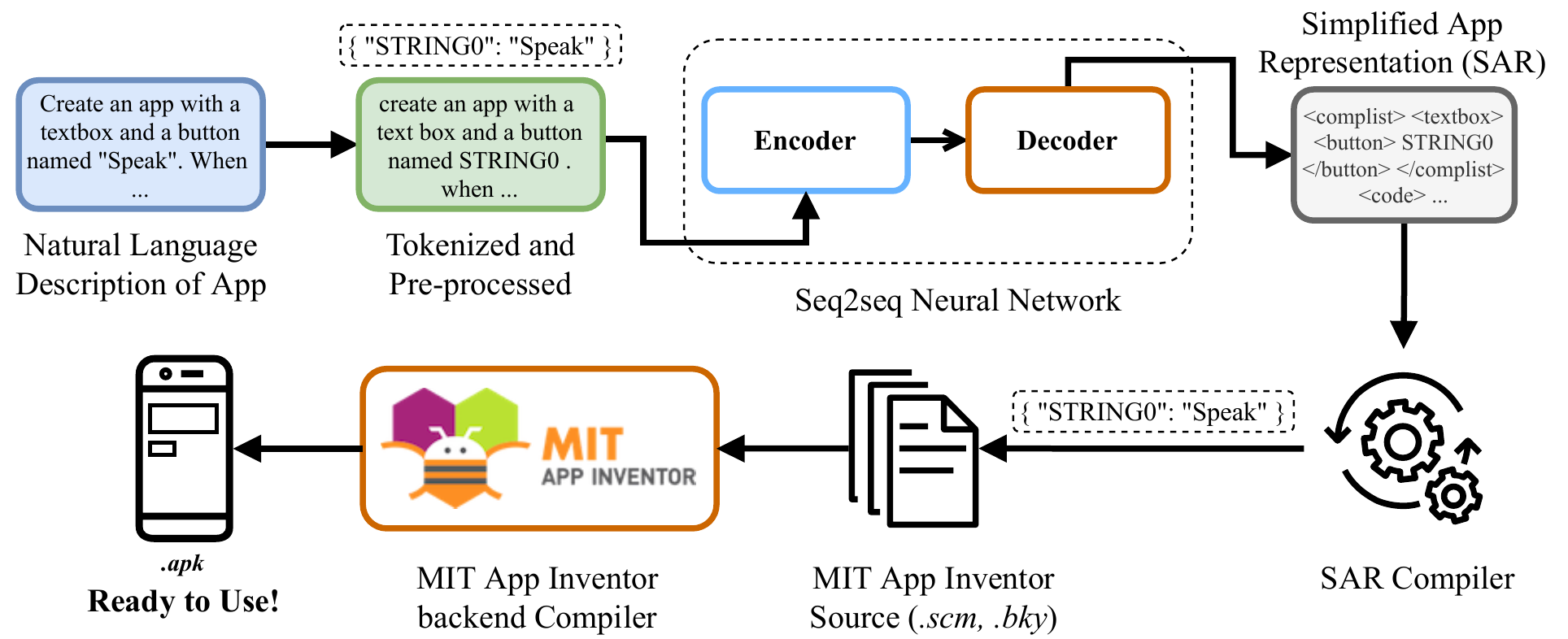}
    \caption{Text2App Prediction Pipeline. A given text is formatted and passed to a seq2seq network to be translated into SAR. Using a SAR Compiler, it is converted to App Inventor project, which can be built into an application.}
    \label{fig:pipeline}
\end{figure*}

Text2App is a framework that aims to build operational mobile applications from natural language (NL) specifications. We make this possible by translating a specification to an intermediate, compact, formal representation which is compiled into the application source code in a later step. This intermediate language helps our system represent an application with a substantially smaller number of tokens, allowing seq2seq models to generate intricate apps in a few decoding steps, which otherwise would be unsolvable by current sequential models.

We design a formal language named \textit{Simplified App Representation (SAR)} that captures the app design, components, and functionalities in a small number of tokens (Section \ref{subsec:sar}). We further develop a SAR compiler that converts a SAR to an application source code (Section \ref{subsec:dsl}). Using MIT App Inventor\footnote{\url{https://appinventor.mit.edu/}} -- a popular, accessible application development tool -- the source code can be compiled to functional app in a matter of minutes. Training a sequence-to-sequence neural network for translating a natural language to SAR requires a parallel NL-SAR corpus. However, human annotation of such a corpus is difficult, and it limits our capability to add new components and functionalities. Instead, we conduct a human survey to understand user perception of text-based app development and app description pattern (Section \ref{subsec:data-collection}), and based on this survey, we create a data synthesis method to automatically generate fluent natural language descriptions of apps along with corresponding SARs (Section \ref{subsec:data-synthesis}). To make sure our synthetic dataset is not monotonous, we introduce a BERT based data augmentation method (Section \ref{subsec:data-augmentation}). Using the synthesized and augmented parallel NL-SAR data, we train multiple sequence-to-sequence neural networks to predict SAR from a given text description of an app, which is then compiled into functional apps (Section \ref{subsec:nmt}). Fig. \ref{fig:pipeline} describes each step in our natural language to app generation process.  We also discuss how a pretrained language model, such as GPT-3, can be used with our system as an external knowledge-base for simplifying abstract human instructions (Section \ref{subsec:gpt-3}). Our system is built to be modular, where each module is self-contained: independent and with a single, well-defined purpose. This allows us to modify one part of the system without affecting the others and debug the system to pinpoint any error.

Literals like strings, numbers are separated during the preprocessing and are re-introduced during compilation. Contrary to conventional programming languages, unless a user specifies a detail of an app component, a suitable default is assumed. This allows the user to describe an app more naturally and also reduces unnecessary overhead from the sequential model.

\subsection{Simplified App Representation (SAR)}
\label{subsec:sar}


SAR is an abstract, intermediate, formal language that represents a mobile application in our system. We design SAR to be minimal and compact, at the same time, to completely describe an application. We formally define the Context Free Grammar of SAR using the following production rules:

\setlength{\grammarparsep}{0pt plus 0pt minus 1pt} 

{\small
\begin{grammar}
<SAR> $\rightarrow{}$ <screens>

<screens> $\rightarrow{}$ <screen> `<NEXT>' <screens> | <screen>

<screen> $\rightarrow{}$ <complist> <code>

<complist> $\rightarrow{}$ `<COMPLIST>' <comps> `</COMPLIST>'

<comps> $\rightarrow{}$ <comps> <comp> | <comp>

<comp> $\rightarrow{}$ `<COMP>' <args> `</COMP>' | `<COMP>'

<code> $\rightarrow{}$ `<CODE>' <events> `</CODE>'

<events> $\rightarrow{}$ <event> <events> | <event>

<event> $\rightarrow{}$ `<EVENT>' <actions> `</EVENT>'

<actions> $\rightarrow{}$ <actions> <action> | <action>

<action> $\rightarrow{}$ `<ACTION>' <args> `</ACTION>'

<args> $\rightarrow{}$ <arg> <args> | <arg>

<arg> $\rightarrow{}$ `<ARG>' `<VAL>' `<ARG>' | `<VAL>'

\end{grammar}
}

Here, \textit{\textless SAR\textgreater} is the starting symbol and the tokens inside quotes are terminals. A mobile application in our system firstly consists of screens. Each screen contains an ordered list of visible (e.g. video player, textbox) or invisible (e.g. accelerometer, text2speech) components, which are identified with the \texttt{<COMPLIST>} tokens. Next, the application logic is defined within the \texttt{<CODE>} tokens.
One functionality in our system is a tuple containing an \textit{event}, an \textit{action}, and a \textit{value}. \texttt{<EVENT>} is an external or internal process that triggers an action. \texttt{<ACTION>} is a process that performs a certain operation. Both \texttt{<EVENT>} and \texttt{<ACTION>} components often have properties that determine their identity or behavior. For example, an animated ball has properties `color', `speed', `radius', etc. Such a property is called an argument (\texttt{<ARG>}). The values of such arguments are indicated by \texttt{<VAL>}. As an example, Figure \ref{fig:app} shows the SAR of an app containing button and text2speech. \texttt{\small <button1\_clicked>} event triggers the action \texttt{\small <speak>} from the text2speech component, which uses \texttt{\small <textbox1text>} - the text in textbox1 as a value.




\subsection{Converting SAR to Mobile Apps}
\label{subsec:dsl}
We convert SAR to MIT App Inventor (MIT AI) project using a custom written compiler. The project is then compiled into functioning app (\textit{.apk}) using MIT AI server. MIT AI is a popular tool for app development large community of active developers, rich and growing functionalities. MIT AI file structure mainly consists of a Scheme (\textit{.scm}) file consisting of components and their properties and a Blockly (\textit{.bky}) file consisting code functionalities. Appendix A shows an algorithm for our SAR to source conversion process. Appendix B and C respectively shows the \textit{.scm} and \textit{.bky} files for the example shown in Fig. \ref{fig:app}. 


The SAR tokens have corresponding predefined template source codes. The compiler parses the tokens and fetches their corresponding templates. By fetching and modifying the predefined templates with the user specified arguments, we generate the \textit{.scm} and \textit{.bky} files from SAR.
The files are then compressed into an MIT AI project file (\textit{.aia}). This has to be uploaded to the publicly available MIT AI server, after which the user can debug the app, or download it as an executable (\textit{.apk}) file.

%



\begin{figure*}
    \centering
    \includegraphics[width=\linewidth]{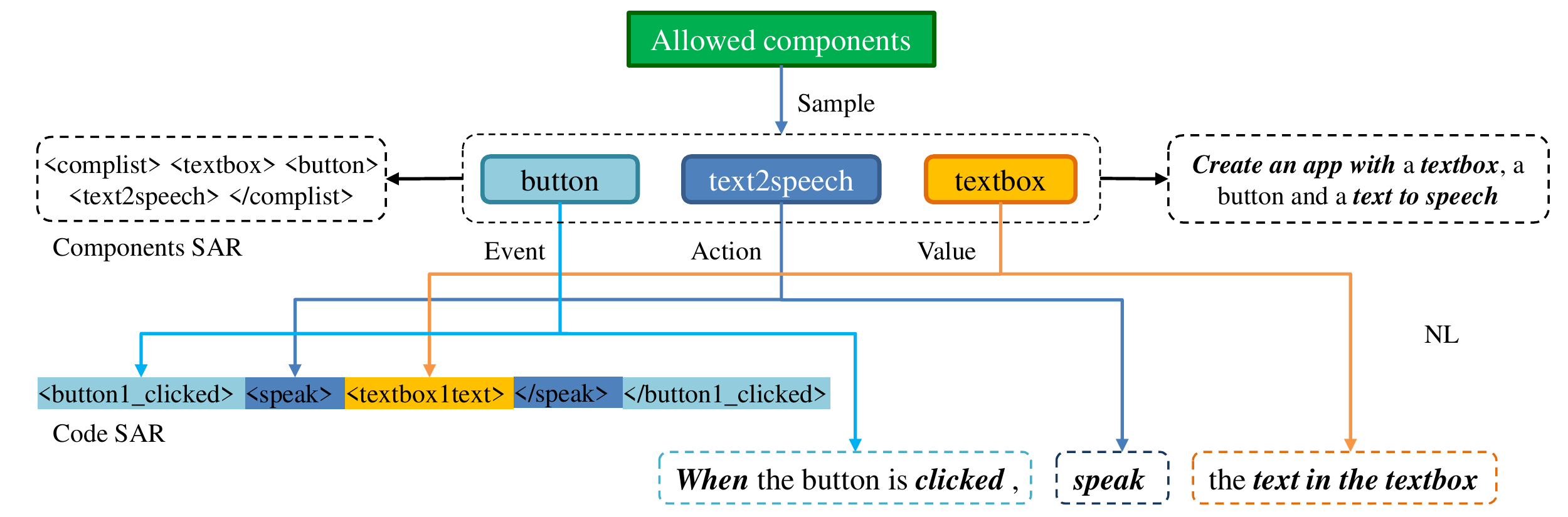}
    \caption{Automatic synthesis of NL and SAR parallel corpus. \textbf{\textit{Bold-italic}} indicates text is selected stochastically.}
    \label{fig:synthesis}
\end{figure*}

\subsection{Survey on Natural Language based App Development}
\label{subsec:data-collection}

In order to understand how a user would perceive a system like Text2App, early in our study, we performed a semi-structured human survey among participants with some programming experience. We asked them to describe several mobile apps from a given set of app components. We received a total of 57 responses\footnote{\url{http://bit.ly/AppDescriptions}} from 30 participants. 36 out of the 57 responses contained enough details for app creation. 19 responses were not detailed, and would require more details and knowledge to be converted to app (e.g. \textit{``make a photo editing app"} -- requires the system to know what a photo editor is).
The observations from this survey helps us to create a data synthesis method, and it works as a general guideline for our study design.

\subsection{Synthesising Natural Language and SAR Parallel Data}
\label{subsec:data-synthesis}

Training a seq2seq model to generate SAR from natural language would require an NL-SAR parallel corpus.
Based on the findings of our survey described in Section \ref{subsec:data-collection}, we develop a data synthesis method for generating natural language app description and SAR data parallelly. First, from a list of allowed components, we randomly select a certain number of components. Most components (i.e. \texttt{button}, \texttt{textbox}) are allowed to be repeated a certain number of times, but some components (e.g. \texttt{text2speech}, \texttt{accelerometer}) can only appear once. The selected components are sorted into three groups, \textit{event component}, \textit{action component}, and \textit{value component} (detailed in Section \ref{subsec:sar}). For each event, action, and value components, their functionalities are selected randomly from a predefined list. When the components, arguments, and their functionalities are selected, we stochastically create a natural language description by sampling natural language snippets from predefined lists. Furthermore, we deterministically create a SAR representation of an app and the functionalities. Figure \ref{fig:synthesis} demonstrates creation of a simple app with three components. The random selection process and repetition of components allows our synthesis method to create wide variety of apps.


\subsection{BERT Based NL Augmentation}
\label{subsec:data-augmentation}

Data augmentation is common practice in computer vision, where an image is rotated, cropped, scaled, or corrupted, in order to increase the data size and introducing variation to the dataset.
To add diversity to our synthetic dataset, we propose a data augmentation method where we mask a certain percentage of words in our dataset using the Masked Language Modeling (MLM) property of pretrained BERT \cite{bert} and sample contextually correct alternate words. Table \ref{tab:augmentation} shows an example of the augmentation technique.




\begin{table}[]
\small
\begin{tabular}{p{7.2cm}}
\toprule
\textbf{\textcolor{blue}{Original:}} Create an app that has an audio player with source string0, a switch. If the switch is flipped, play player. \\ \midrule
\textbf{\textcolor{blue}{Augmentation:}} Create an app that has an \colorbox{green}{external} player with source string0, a switch. If the switch \colorbox{green}{gets} flipped, play player. \\ \bottomrule
\end{tabular}
\caption{BERT mask filling based data augmentation method. Mutated words are highlighted green.}
\label{tab:augmentation}
\end{table}

\subsection{NL to SAR Translation using Seq2Seq Networks}
\label{subsec:nmt}
We generate 50,000 unique NL and SAR parallel data using our data synthesis method, and mutate 1\% of the natural language tokens. We split this dataset into train-validation-test sets in 8:1:1 ratio and train three different models.

\noindent
\textbf{Pointer Network:} We train a Pointer Network \cite{pt-gen} consisting of a randomly initialized bidirectional LSTM encoder with hidden layers of size 500 and 250.

\noindent
\textbf{Transformer with pretrained encoders:} We create two sequence-to-sequence Transformer \cite{transformer} networks each having 12 encoder layers and 6 decoder layers. Every layer has 12 self-attention heads of size 64. The hidden dimension is 768. 
The encoder of one of the models is initialized with RoBERTa base \cite{roberta} pretrained weights, and the other one with CodeBERT base \cite{codebert} pretrained weights. 

\subsection{Simplifying Abstract Natural Language Instructions using GPT-3}
\label{subsec:gpt-3}

\input{Tables/results}

\begin{table}[h]
\small
\begin{tabular}{p{7.2cm}}
\toprule
1. \textbf{number adding app -} make an app with a textbox, a textbox, and a button named "+". \\
\textbf{SAR: } $<$complist$>$ $<$textbox$>$ $<$textbox$>$ $<$button$>$ + $<$/button$>$ $<$/complist$>$
\\ \midrule
2. \textbf{twitter app -} make an app with a textbox, a button named ``tweet", and a label. When the button is pressed, set the label to textbox text.\\
\textbf{SAR:} $<$complist$>$ $<$textbox$>$ $<$button$>$ tweet  $<$/button$>$ $<$label$>$ label1 $<$/label$>$ $<$/complist$>$ $<$code$>$ $<$button1clicked$>$ $<$label1$>$ $<$textboxtext1$>$ $<$/label1$>$ $<$/button1clicked$>$ $<$/code$>$
\\ \midrule
3. \textbf{browser app -}  create an app with a textbox, a button named ``go", and a button named ``back". When the button ``go" is pressed, \textcolor{red}{\textit{go to the url}} in the textbox. When the button ``back" is pressed, go back to the \textcolor{red}{\textit{previous page}}. \\ \midrule
4. \textbf{Google front page -} make an app with a textbox, a button named ``google", and a button named ``search". When the button ``google" is pressed, \textcolor{red}{\textit{search google}}. When the button ``search" is pressed, \textcolor{red}{\textit{search the web}}. \\ \midrule
\end{tabular}
\caption{Abstract instructions to simpler app description using GPT-3. Example 1, 2, was constrained within our allowed set of functionalities. 3, 4, introduced concepts that are not yet supported in Text2App (marked in \textcolor{red}{\textit{red and italic}}).}
\label{tab:simplification}
\end{table}


In our survey sessions (Section \ref{subsec:data-collection}) we found that some abstract instructions require external knowledge to be converted to applications (e.g. \textit{``Create a photo editor app"} -- expects knowledge how a photo editor looks and works). Large pretrained autoregressive language models (LMs), have shown to understand abstract natural language concepts and even explain them in simple terms \cite{gpt3-explains}. Using the few shot prediction capability of GPT-3 \cite{gpt-3}, we experiment with simplifying abstract app concepts. We find that although this method is promising, the LM fails to limit the prediction to our current capability (Table \ref{tab:simplification}). The GPT-3 prompts are provided in Appendix D.


%% file: Tables/results.tex
\begin{table*}[!htb]
\centering
\resizebox{0.9\textwidth}{!}{
\begin{tabular}{llcccccccccc}
\toprule
\multicolumn{1}{c}{\multirow{3}{*}{}} & \multirow{3}{*}{\textbf{\#Epoch}} & \multicolumn{2}{c}{\multirow{2}{*}{\textbf{Test}}} & \multicolumn{6}{c}{\textbf{BERT Mutation}} & \multicolumn{2}{c}{\multirow{2}{*}{\textbf{Unseen Pair}}} \\ \cline{5-10}
 &  & \multicolumn{2}{c}{} & \multicolumn{2}{c}{\textbf{2\%}} & \multicolumn{2}{c}{\textbf{5\%}} & \multicolumn{2}{c}{\textbf{10\%}} & \multicolumn{2}{c}{} \\ \cline{3-12} 
\multicolumn{1}{c}{} &  & \multicolumn{1}{c}{\textbf{BLEU}} & \multicolumn{1}{c}{\textbf{EM}} & \multicolumn{1}{c}{\textbf{BLEU}} & \multicolumn{1}{c}{\textbf{EM}} & \multicolumn{1}{c}{\textbf{BLEU}} & \multicolumn{1}{c}{\textbf{EM}} & \multicolumn{1}{c}{\textbf{BLEU}} & \multicolumn{1}{c}{\textbf{EM}} & \multicolumn{1}{c}{\textbf{BLEU}} & \multicolumn{1}{c}{\textbf{EM}} \\ 
\midrule
\multicolumn{12}{c}{\textbf{Without Training Data Augmentation}} \\
\midrule
\textbf{PointerNet} & 13.6 & 94.64 & 79.24 & 94.16 & 72.06 & 91.80 & 56.14 & 88.96 & 40.78 & 96.75 & 82.91 \\ 
\textbf{RoBERTa init} & 3 & 97.20  & 77.80 & 96.83 & 73.20 & 94.97 & 61.68 & 92.86  & 48.06 & 98.11  & 79.66 \\ 
\textbf{CodeBERT init} & 8 & 97.42 & 80.02 & 97.18 & 76.02 & 95.37 & 64.38 & 93.29 & 51.24 & 98.47  & 83.50 \\
 \midrule
\multicolumn{12}{c}{\textbf{With  Training Data Augmentation (1\% Mutation)}} \\
\midrule
\textbf{PointerNet} & 23.2 & 95.03 & 81.40 & 94.85 & 79.46 & 93.85 & 72.04 & 92.53 & 63.68 & 96.68 & 83.33 \\ 
\textbf{RoBERTa init} & 3 & \textbf{97.66} & \textbf{81.76} & \textbf{97.60} & \textbf{80.66} & \textbf{96.91} & \textbf{76.16} & \textbf{96.04} & \textbf{70.10} & \textbf{98.64} & \textbf{84.68} \\ 
\textbf{CodeBERT init} & 7 & 97.64 & 81.66 & 97.51 & 80.20 & 96.74 & 74.98 & 95.71 & 67.58 & 98.62  & 84.51 \\ 
\bottomrule
\end{tabular}
}
\caption{Comparison between Pointer Network and seq2seq Transformer with encoder initialized with RoBERTa and CodeBERT pretrained weights. BLEU indicates BLEU-1 and Exact Match (EM) is shown in percent.}
\label{tab:model-results}
\end{table*}

%% file: Sections/Evaluation.tex
\textbf{Automatic Evaluation.}
We evaluate the three seq2seq networks mentioned in Section \ref{subsec:nmt} -- PointerNetwork, seq2seq Transformer initialized with RoBERTa, and seq2seq Transformer initialized with CodeBERT.
We evaluate the models in 3 different settings -- firstly, in a held out test set, secondly, with increasing amount of mutation in the test set (2\%, 5\% 10\%) (Section \ref{subsec:data-augmentation}). We also trained separate models using data excluding specific combinations of components (\texttt{\small <button1clicked>, <text2speech>}) and then tested them on the excluded data. These establishes the models' ability to generalize beyond the patterns it was trained on. 
From Table \ref{tab:model-results} we can see that augmenting the training dataset notably improves all models (up to 22.04\%). We also see that the RoBERTa initialized model performs best in all evaluation categories. Note that, all predictions reported in Table \ref{tab:model-results} are valid SAR format.


\textbf{Human Evaluation.}
To evaluate our system with real world natural language, we conduct a survey with 13 Computer Science undergraduate volunteers. We provided the participants short videos of 10 mobile applications, and asked them to describe the application in their own language\footnote{\url{https://bit.ly/text2appsurvey}}. We provided them with the component names and one example NL. We collected total 112 responses and generated the SAR from these responses using Text2App system. The evaluation contained application SARs with different lengths (min 10, max 42) Upon comparing the generated SAR with the ground truth SAR data we found a BLEU-1 score of 54.99, and exact match 4/111. The labeled data and predictions are made publicly available\footnote{\url{https://bit.ly/text2appsurveyresponses}}. Upon manual inspection, we found that 25 out of the 112 predictions are either correct or functionally correct predictions of the user provided NL. This represents the difficulty of our task and the complexity of natural language. Observing the successful and unsuccessful predictions, we find the patterns emerge as shown in Table \ref{tab:observation}.


\begin{table}[h]
\small
    \centering
    \begin{tabular}{p{2.9cm}|p{3.8cm}}
        \toprule
         \textbf{Observation} & \textbf{Example}  \\
        \midrule
         Successful predictions closely follow our data format. (i.e. clear component list followed by functionality.) & \textbf{NL:} \textit{Create an app that has a button named ``Take Photo", when clicked open the rear camera and capture an image.} \\
         \midrule
         Many predictions are correct, but the human label is wrong. & \textbf{NL:} \textit{Create an app that can take a photo with the camera. } (Missing mention of a button.)\\
         \midrule
         User expects the automatic system to have inherent understanding of the real world. & \textbf{NL:} \textit{A typical registration form with necessary text fields and a submit button.} \\
         \midrule
         Model is biased towards synthesizer keywords. & \textbf{NL:} \textit{Create an app which has a moving ball.} (`moving' frequently appears with accelerometer, model predicts accelerometer component.)\\
         \midrule
         System misses components not specified. & \textbf{NL:} \textit{Create an app that has a textbox labelled "Insert Text", when the device is shaken, speak the text in the textbox} \\
        \bottomrule
    \end{tabular}
    \caption{Observation from human labeled data predictions}
    \label{tab:observation}
\end{table}

%% file: Sections/DiscussionAndFuture.tex
Text2App is the first attempt of an NL based app development tool, and our core contribution of our project lies in the development endeavour of building the SAR, the SAR compiler, and the SAR-NL parallel data synthesizer. Currently it supports app creation from 12 components (i.e. \textit{`camera', `textbox', `button', `text2speech', `ball', `accelerometer', `video_player', `switch', `player', `label', `timepicker', `passwordtextbox'}), 4 events (i.e. \textit{button click, switch flip, accelerometer shaken, ball flung}), and more than 10 actions (e.g. \textit{Take a picture, Start/stop video/audio, Speak a given text or a textbox, Bounce ball, set speed/color of the ball, Set label, etc.}). We have covered the first 3 out of 4 fundamentals of Android applications (i.e. Activity, Services, Content Provider, Broadcast Receiver). Most components supported by in MIT App Inventor follows a similar tree-like pattern, and thus can be represented as SAR. We are inviting open source community to contribute to and help grow this project. 
In future, we would like to experiment with generative model based data synthesis, more reliable language model based NL simplification techniques, and potentially app development in native languages. 

%% file: Sections/Conclusion.tex


In this paper, we explore creating functional mobile applications from natural language text descriptions using seq2seq networks. We propose Text2App, a novel framework for natural language to app translation with the help of a simpler intermediate representation of the application. The intermediate formal representation allows to describe an app with significantly smaller number of tokens than native app development languages. We also design a data synthesis method guided by a human survey, that automatically generates fluent natural language app descriptions and their formal representations. Our AI aware design approach for a formal language can guide future programming language and frameworks development, where further source code generation works can benefit from.

%% file: Sections/Acknowledgement.tex
We thank Prof. Zhijia Zhao from UCR for proposing the problem that inspired this project idea.

We also thank OpenAI, Google Colaboratory, Hugging Face, MIT App Inventor community, the survey participants, and Prof. Anindya Iqbal for feedback regarding modularity. This project was funded under the `Innovation Fund' by the ICT Division, Government of the People’s Republic of Bangladesh.

%% file: Sections/AppendixText.tex




\section{SAR Compilation Algorithm}

{\small
\begin{algorithm}[h] 
\DontPrintSemicolon

  \textbf{Input} SAR tokens, LiteralDict
  
  \textbf{Output} .scm, .bky
  
    scm = initializeSCM() bky = initializeBKY()
    
    \For{token in complist}{
    	\If{isComponentStart(token)}{
    		uuid = generateNumUUID()
    		
    		n = getCompNum(token)
    		
    		\If{token.hasArgument()}{
    			args = fetchArgs(token, LiteralDict)}
    			
    		t = getTemplate(token)
    		
    		t.set(n, uuid, args)
    	}
    	scm.add(t)
    	
    }
    write(scm)
    
    \For{token in code}{
        t = getTemplate(token)
        
        uuid = generateStringUUID()
        
        \If{token.isLiteral()}{
            val = LiteralDict[token] 
            
            \If{val.isFileDir() \& fileDoesNotExist}{
                val = closestMatchingFile()
            }
            t.set(val)
        }
        \Else{
            \If{token.hasNumber()}{
                number = regexMatch(token)
                
                t.set(number)
            }
        }
        t.set(uuid)
        
        bky.add(t)
    }
    write(bky)
\caption{Compiling SAR to \textit{.scm}, \textit{.bky}}
\label{alg:sar-compiler}
\end{algorithm}
}

\section{Sceme (\textit{.scm}) File for Visual Components}

\lstset{frame=tb,
  showstringspaces=false,
  columns=flexible,
  basicstyle={\small\ttfamily},
  numbers=none,
  numberstyle=\tiny\color{black},
  keywordstyle=\color{black},
  commentstyle=\color{black},
  stringstyle=\color{black},
  breaklines=true,
  breakatwhitespace=true,
  tabsize=4
}

\begin{lstlisting}[breaklines, showstringspaces=false, escapeinside={(*@}{@*)}, caption={A sample scm file representing the visual components of the app in Fig. 1. The blue portion lists the components of the app.},captionpos=b, label={lst:scm}]
#|
$JSON
{"authURL":["ai2.appinventor.mit.edu"], 
"YaVersion":"208",
"Source":"Form",
"Properties":{"$Name":"Screen1","$Type": "Form","$Version":"27",
"AppName":"speak_it","Title":"Screen1", "Uuid":"0",(*@
\newline\textcolor{blue}{"\$Components":[\{"\$Name":"TextBox1", "\$Type":"TextBox","\$Version":"6", \newline "Hint":"Hint for TextBox1","Uuid":"913409813"\},\{"\$Name": "Button1", \newline "\$Type":"Button", "\$Version":"6","Text":"Speak","Uuid": "955068562"\}, \newline \{"\$Name":"TextToSpeech1", "\$Type":"TextToSpeech","\$Version":"5", \newline "Uuid":"1305598760"\}]}@*)}}
|#
\end{lstlisting}

\section{Blockly (\textit{.bky}) Logical Components}

\begin{lstlisting}[breaklines, showstringspaces=false, escapeinside={(*@}{@*)}, caption={A sample bky file representing the logical components of the app in Fig. 1. The colored lines represent different blocks.},captionpos=b, label={lst:bky}]
<xml xmlns="http://www.w3.org/1999/xhtml">
  (*@ \textcolor{orange}{<block type="component\_event" id="gnc7Dj5so`[8HB\}z|Ohk" x="-184" y="91">}@*)
    (*@\textcolor{orange}{<mutation component\_type="Button" is\_generic="false" instance\_name="Button1" event\_name="Click"></mutation>}@*) 
    (*@\textcolor{orange}{<field name="COMPONENT\_SELECTOR"> Button1 </field>}@*) 
    (*@\textcolor{orange}{<statement name="DO">}@*)
      (*@\textcolor{violet}{<block type="component\_method" id="-7*:E7Xk@uO5?b32/Gq3">}@*)
        (*@\textcolor{violet}{<mutation component\_type="TextToSpeech" method\_name="Speak" is\_generic="false" instance\_name="TextToSpeech1"> </mutation>}@*)
        (*@\textcolor{violet}{<field name="COMPONENT\_SELECTOR"> TextToSpeech1 </field>}@*)
        (*@\textcolor{violet}{<value name="ARG0">}@*)
          (*@\textcolor{green}{<block type="component\_set\_get" id="wS:Fm\{EYxQ]B1\%*LO2zp">}@*)
            (*@\textcolor{green}{<mutation component\_type="TextBox" set\_or\_get="get" property\_name="Text" is\_generic="false" instance\_name="TextBox1"> </mutation>}@*)
            (*@\textcolor{green}{<field name="COMPONENT\_SELECTOR"> TextBox1 </field>}@*)
            (*@\textcolor{green}{<field name="PROP">Text</field>}@*)
          (*@\textcolor{green}{</block>}@*)
        (*@\textcolor{violet}{</value>}@*)
      (*@\textcolor{violet}{</block>}@*)
    (*@\textcolor{orange}{</statement>}@*)
  (*@\textcolor{orange}{</block>}@*)
  <yacodeblocks ya-version="208" language-version="33"></yacodeblocks>
</xml>
\end{lstlisting}

\clearpage
\section{GPT-3 Prompt}

\begin{table}[h]
\small
\begin{tabular}{p{7.2cm}}
\toprule
How to make an app with these components : button, switch, textbox, accelerometer, audio player, video player, text2speech

\textcolor{brown}{random video player app.} -- make an app with a video player with a random video, a button named ``play" and a button named ``pause". When the first button is pressed, start the video. When the second button is pressed pause the video.

\textcolor{brown}{A time speaking app} -- an app with a button, a clock and a text2speech. When the button is clicked, speak the time.

\textcolor{brown}{Display time app} -- create an app with a button, a timepicker, and a label. When the button is pressed, set the label to the time.

\textcolor{brown}{A messeging app} -- create an app with a with a textbox, and a button named "send", and a label. When the button is pressed, set label to textbox text.

\textcolor{brown}{Login form} -- create an app with a textbox, a passwordbox, and a button named "login".

\textcolor{brown}{Search interface} -- make an application with a textbox, and a button named ``search".

\textcolor{brown}{siren app} -- create an app with a music player with source ``siren\_sound.mp3", and a button. When the button is pressed, play the audio.

\textcolor{brown}{An arithmatic addition app gui} -- make an app with a textbox, a textbox, and a button named ``+".

\textcolor{brown}{vibration alert app} -- create an app with an accelerometer, and a text2speech. When the accelerometer is shaken, speak ``vibration detected".

\textcolor{brown}{\textit{\{A new prompt\}}} -- \\
\bottomrule
\end{tabular}
\caption{Prompt used to generate app description with GPT-3. A new unseen prompt is added at the end and the model is tasked to continue generating text in the same pattern. This method is known as Few Shot text generation \cite{gpt-3}.}
\label{tab:simplification2}
\end{table}